\documentclass[11pt, a4paper, logo, copyright, nonumbering]{deepseek}
\usepackage[authoryear, sort&compress, round]{natbib}
\usepackage{ulem}
\usepackage{caption}
\usepackage{xspace}
\usepackage{pifont}
\usepackage{multirow}
\usepackage{tcolorbox}
\usepackage{xltabular}
\usepackage{longtable}
\usepackage{hyperref}
\usepackage[capitalize,noabbrev]{cleveref} 
\usepackage{lipsum}  
\usepackage{multicol} 
\usepackage{amsfonts}
\usepackage{amsmath}
\usepackage{amssymb}
\usepackage{lineno}

\usepackage{mathtools}
\usepackage{mathrsfs}
\usepackage{physics}

\usepackage[bottom]{footmisc}

\usepackage{subcaption}
\usepackage{xcolor}

\renewcommand{\today}{}

\title{\centering Echo: Learning from Experience Data via User-Driven Refinement}

\author{
    \vspace{-0.6cm}
    Hande Dong\protect\footnotemark[1], 
    Xiaoyun Liang\protect\footnotemark[1], 
    Jiarui Yu\protect\footnotemark[1], 
    Jiayi Lin\protect\footnotemark[1], 
    Changqing Ai\protect\footnotemark[1], 
    Feng Liu\protect\footnotemark[1], 
    Wenjun Zhang\protect\footnotemark[1], 
    Rongbi Wei, 
    Chaofan Zhu, 
    Linjie Che, 
    Feng Wu, 
    Xin Shen,
    Dexu Kong,
    Xiaotian Wang,
    Qiuyuan Chen,
    Bingxu An,
    Yueting Lei,
    Qiang Lin\protect\footnotemark[2] \\
    
    \vspace{-1cm}
}

\begin{document}

\begin{abstract}

Static "human data" faces inherent limitations: it is expensive to scale and bounded by the knowledge of its creators. Continuous learning from "experience data"—interactions between agents and their environments—promises to transcend these barriers. Today, the widespread deployment of AI agents grants us low-cost access to massive streams of such real-world experience. However, raw interaction logs are inherently noisy, filled with trial-and-error and low information density, rendering them inefficient for direct model training.

We introduce \textbf{Echo}, a generalized framework designed to operationalize the transition from raw experience to learnable knowledge, effectively "echoing" environmental feedback back into the training loop for model optimization. 
In today's agent ecosystem, user refinement serves as a primary source of such feedback: driven by responsibility for the outcome, users rigorously transform flawed agent proposals into verified solutions. 
These user-driven refinement sequences inherently distill agents' crude attempts into high-quality training signals. Echo systematically harvests these signals to continuously align the agent with real-world needs. Large-scale validation in a production code completion environment confirms that Echo effectively harnesses this pipeline, breaking the static performance ceiling by increasing the acceptance rate from 25.7\% to 35.7\%.


\end{abstract}

\maketitle

\begin{figure}[h!]
    \centering
    \includegraphics[width=0.95\textwidth]{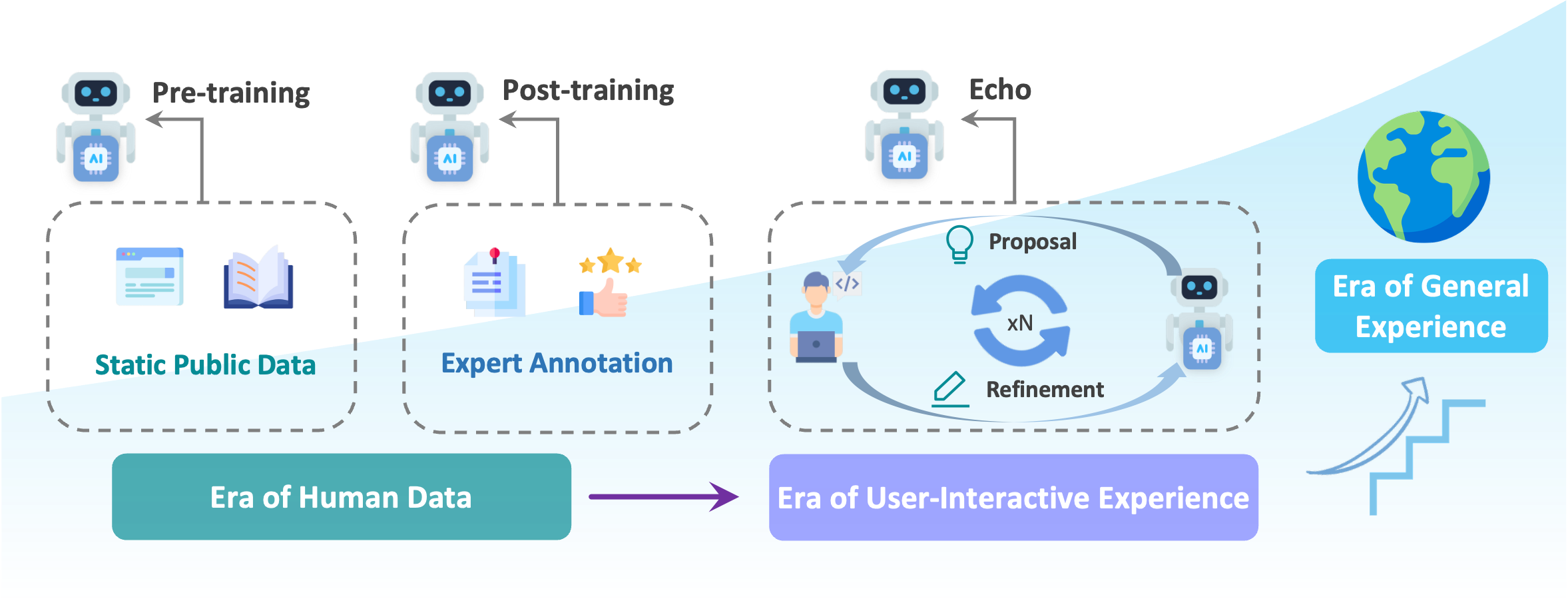}
    \caption{\textbf{Paradigm Shift: From Static Archives to Interactive Experience.} Traditional training relies on finite human data and expert annotations (left). The Echo framework (center) breaks this static ceiling by harvesting continuous, user-driven refinements. Distilling these user corrections scales the agent toward the open-ended Era of General Experience (right).}
    \label{fig:placeholder}
\end{figure}

\footnotetext[1]{Core Contributors. Hande Dong is the project leader (\texttt{donghd66@gmail.com}).}
\footnotetext[2]{Qiang Lin is the team leader.}

\section{Introduction}

Current paradigms in agent training are largely predicated on the success of massive, static "human data." This approach—comprising pre-training, supervised fine-tuning (SFT), and reinforcement learning—relies on archived web content and expert annotationss~\citep{ouyang2022training, rafailov2023direct, chaudhari2025rlhf}. However, as agents strive for superhuman capabilities, this paradigm is increasingly challenged by the finite nature of high-quality human data~\citep{villalobos2024position}. More importantly, this limits agent intelligence strictly within the distribution of its fixed training corpus, making performance gains progressively difficult to achieve~\citep{ross2011reduction, shumailov2024ai}.

To break these boundaries, the vision of "experience data" has emerged—proposing that agents should learn from their continuous interactions with the environment~\citep{yao2022react, shridhar2020alfworld, koh2024visualwebarena, zhang2025landscape, dulac2019challenges, pan2025measuring}. Today, the widespread deployment of AI agents (such as Cursor, Perplexity, and Claude Code) grants us low-cost access to massive streams of such real-world experience~\citep{bakal2025experience, han2025reinforcement, cui2025effects, knox2008tamer, christiano2017deep, chen2024learning}. However, raw interaction logs are inherently noisy, filled with trial-and-error and low information density, rendering them inefficient for direct model training~\citep{xie2024osworld, zhou2023webarena, silver2025welcome}. The core challenge is mining valuable, high-fidelity training signals from this chaotic open-world noise~\citep{m2023model}.

To operationalize this paradigm shift, we introduce Echo, a methodology designed to extract learnable knowledge from raw experience. We formalize this evolution through three core conceptual pillars: 
\begin{itemize}[leftmargin=*]
    \item \textbf{Experience Acquisition}: We define the continuous, streamful interaction between the agent and its real-world environment as its raw experience. While capturing the unvarnished reality of problem-solving, these captured streams are inherently noisy and suffer from low information density, making them inadequate for direct model training.
    \item \textbf{Knowledge Extraction}: To bridge this gap, we formalize the transition from experience to knowledge based on a core hypothesis: in outcome-oriented agent applications, users inherently act as accountable stakeholders~\citep{amershi2019guidelines}. Driven by this accountability, they actively drive flawed agent proposals to the final correct outcome—a mechanism we formalize as user-driven refinement. Through this active refinement, users naturally embed the missing knowledge into the solution. The resulting outcomes thus encapsulate high-entropy, compressed signals that are highly effective for model training. 
    \item \textbf{Model Optimization}: Finally, we formulate the optimization objective. Rather than modeling the noisy intermediate paths of human editing, the agent learns to directly align its generative distribution with these refined outcomes, enabling autonomous evolution toward the correct destination.
\end{itemize}

To evaluate the effectiveness of this framework, we implemented Echo within the code completion scenario, extracting experience data from the massive interaction streams of developers working at Tencent Cloud. The results provide a decisive proof of concept: Echo broke the performance ceiling of static baselines, increasing the code acceptance rate from 25.7\% to 35.7\%. Beyond these immediate gains, our analysis reveals two critical implications: first, the model demonstrates robust generalization to external users, indicating it learned underlying logic rather than overfitting to specific patterns; second, we observe a distinct scaling effect where performance improves continuously with data volume without saturation. This confirms that harvesting open-world experience serves as a sustainable engine for perpetual agent learning.

In summary, our contributions are as follows: 
\begin{itemize}[leftmargin=*]
\item \textbf{Conceptual Paradigm Shift}: We formalize the transition from static, archival training to interactive, experience-based learning. We identify user-driven refinement as the critical mechanism that naturally embeds missing knowledge into the final solution.
\item \textbf{The Echo Framework}: We introduce a rigorous, three-stage methodology (Experience $\to$ Knowledge $\to$ Optimization) to operationalize this paradigm, providing a scalable pipeline for extracting high-entropy, compressed training signals from chaotic real-world logs.
\item \textbf{Industrial-Scale Validation}: We deploy and evaluate Echo in a production code completion ecosystem. By directly aligning the model with human-refined outcomes, we demonstrate a significant absolute increase of 10\% in the online acceptance rate, proving the framework's effectiveness in breaking static performance ceilings.
\end{itemize}

\section{Related Work}

The development of Echo is situated at the intersection of large language model (LLM), reinforcement learning (RL), and the emerging paradigm of experience learning.

\subsection{Learning from Static Archives}

The prevailing paradigm for training autonomous agents relies heavily on static "human data" through pre-training and post-training pipelines~\citep{ouyang2022training, rafailov2023direct, chaudhari2025rlhf}. While effective, this approach faces a fundamental ceiling. As agents approach expert-level performance, they inevitably outstrip the fixed distribution of their training corpora, limiting them to the capabilities of their demonstrators~\citep{ross2011reduction}. Furthermore, acquiring supervision that exceeds the agent's own capabilities becomes exponentially expensive, transforming model scaling into a bottleneck of data scarcity~\citep{burns2023weak, villalobos2024position}. These constraints, compounded by the risks of model collapse from ungrounded recursive training~\citep{shumailov2024ai}, necessitate a shift from imitating finite historical archives to learning from "experience data"---evolving through the active consequences of real-world interaction.

\subsection{Reinforcement Learning}

Reinforcement Learning (RL) has become a cornerstone for aligning and enhancing language models. RLHF~\citep{ouyang2022training, christiano2017deep} established the standard by optimizing models against learned reward functions derived from human preferences. Moreover, systems like DeepSeek-R1~\citep{guo2025deepseek} have pushed this frontier, demonstrating that RL can drive profound improvements in reasoning capabilities through large-scale reinforcement on verifiable tasks. 

Despite these successes, a critical gap remains in open-ended domains. Standard RL signals are typically evaluative (scalar rewards) rather than instructive. They inform the agent that a response is preferred, but lack the dense information to explain why or how to correct specific reasoning flaws. This sparsity often leads to sample inefficiency and reward hacking~\citep{casper2023open}, limiting the ability to learn complex, tacit knowledge from real-world interactions.

\subsection{Experience Learning}

Recently, some research explores acquiring experience through synthesis or environmental interaction. DreamGym~\citep{chen2025scaling} employs reasoning-based experience models to distill environment dynamics, synthesizing diverse tasks and feedback to enable scalable online training. 
Shifting toward environmental interaction, Early Experience~\citep{zhang2025agent} relies on multiple offline rollouts to explore environment, utilizing strategies like implicit world modeling and self-reflection to ground policies without relying on rewards. 
In contrast, OEL~\citep{ye2026online} focuses on single-time experiences within online services; it extracts knowledge from such interactions and consolidates this information into model parameters via on-policy context distillation. 

Yet, synthesis is inherently capped by the fidelity of the internal world model, while environmental interaction often lacks authoritative external grounding to verify task success.
Yet, these approaches face significant scaling bottlenecks. Synthesis is inherently capped by the fidelity of the internal world model. Meanwhile, environmental interaction either remains confined to toy-like text games or lacks authoritative external grounding, not being tailored to assimilate experiential knowledge from complex industrial-scale feedback.
Echo bridges this gap by shifting the source of experience from synthetic simulations to authentic, production-level applications. Instead of relying on sparse rewards or internal guessing, Echo leverages accountable stakeholders who actively drive flawed proposals to the final correct outcome. This user-driven refinement naturally embeds the missing knowledge, providing a dense, high-fidelity training target that drives perpetual evolution within the real world.

\section{Echo: Learning from Experience via User-Driven Refinement}

\begin{figure}[t]
    \centering
    \includegraphics[width=0.95\textwidth]{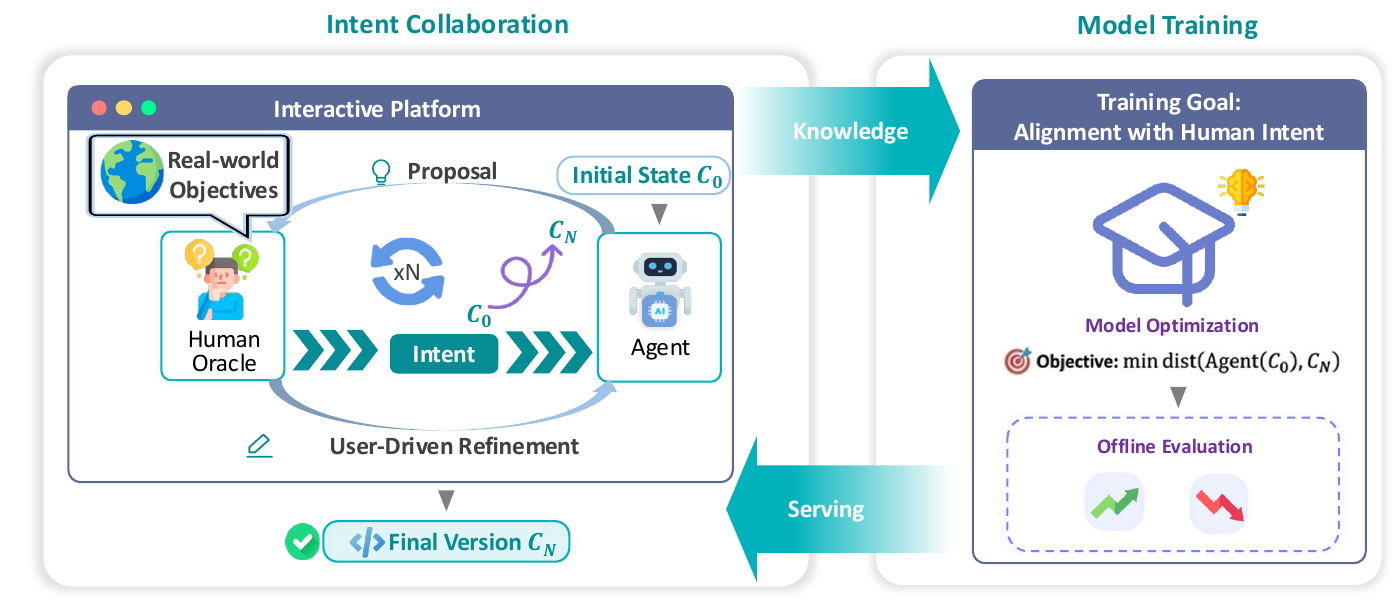}
    \caption{\textbf{Overview of the Echo Framework.} Echo establishes a closed-loop system to transform real-world corrections into agent intelligence. \textbf{User-Driven Refinement}: in daily interaction process, the system captures the discrepancy between the agent's initial proposal ($C_0$) and the user's final commit ($C_N$). \textbf{Model Training:} the refined experience data drives continuous model optimization, enabling the agent to learn from its own real-world deployment iteratively.}
    \label{fig:overview}
\end{figure}

We formalize Echo as a generalized framework grounded in the interactive paradigm of outcome-oriented agentic applications. This framework is inherently environment-agnostic, representing a universal paradigm for agentic learning. Whether an agent is operating a web browser, executing complex multi-step reasoning, or assisting in specialized content creation, the underlying dynamic remains the same: an accountable stakeholder acts as the ultimate environmental grounding, actively driving flawed proposals to the final correct outcome.

To operationalize this universal dynamic, this section breaks down the Echo framework into three core theoretical components:
\begin{itemize}
    \item \textbf{Experience Acquisition}: We define the sequential loop, through which raw, noisy experience is acquired during the agent's service process.
    \item \textbf{Knowledge Extraction}: We then describe the mechanism of knowledge extraction, detailing how human responsibility actively shapes flawed proposals into verified ground truths.
    \item \textbf{Model Optimization}: Finally, we formulate the optimization objective, defining how the agent internalizes these correction trajectories to continuously learn from its lived experience.
\end{itemize}

\subsection{Experience Acquisition through the Agent-Environment Interaction}

Today, AI agents are actively deployed in open-world environments to execute complex tasks. In this setting, the agent and the environment (e.g., a digital workspace) are coupled in a sequential interaction loop. At each step, the environment presents an observable state, denoted as the initial context $C_0$. The agent perceives this state and executes an action $a$ (e.g., generating a proposal) that is committed to the environment and triggers a state transition. This continuous cycle of State ($C_0$) $\to$ Action ($a$) $\to$ Transition constitutes the raw stream of experience data.

However, learning from this raw experience presents a fundamental challenge: the absence of an automatic oracle. In an open-world environment, there is no automatic mechanism to evaluate whether the state transition triggered by the agent's action satisfies the true task objectives. The agent acquires vast amounts of experience through interaction, but it lacks a programmatic signal to judge the correctness of its impact on the environment.

\subsection{Knowledge Extraction via User-Driven Refinement}

To extract learnable signals from raw experience data, Echo formalizes the user as the indispensable accountable stakeholder within the environment, characterizing their core value through user-driven refinement. Driven by responsibility for the final outcome, the professional user actively steers the transformation of the environment toward correctness, achieving this either by driving the agent's intervention or through direct intervention to ensure the evolving artifact strictly adheres to real-world constraints.

Specifically, when the agent automatically perceives the initial environmental state $C_0$ (the context) and generates an initial proposal (denoted as $C_1$, which may be the culmination of a multi-step action sequence), the user evaluates it against actual real-world objectives. If the proposal deviates from these objectives, the user initiates a corrective intervention. This triggers the refinement process, evolving the flawed proposal into a verified solution through a dynamic sequence of user interventions, which can be formally represented as:

$$\text{State } C_0 \xrightarrow{\text{Agent}} \underbrace{C_1 \xrightarrow{\text{User Interventions}} C_N}_{\textbf{Refinement Sequence}}$$

Echo identifies the final state $C_N$ as the compressed knowledge provided by the accountable stakeholder. Through this active refinement, the user naturally injects the knowledge missing from the agent—such as domain-specific context or tacit constraints—into the final artifact. This state represents the ground-truth target for the initial context $C_0$, embodying the correct transformation logic that the agent should have ideally achieved to satisfy the real-world objectives.

We note that resolving $C_N$ from the noisy, streamful interaction data is a non-trivial data mining challenge. While Section 4 will detail a large-scale engineering pipeline as a concrete example of industrial practice, the theoretical framework of Echo remains mathematically agnostic to any specific mining implementation.

\subsection{Model Optimization: Aligning with Compressed Knowledge}

The ultimate objective of Echo is to enable the agent to generate the ground-truth state $C_N$ directly from the initial context $C_0$. In this framework, $C_1$ is recognized as an unsuccessful attempt that failed to meet real-world constraints, while $C_N$ serves as the gold standard for correctness.

By leveraging the accountable stakeholder, Echo treats the verified target state ($C_N$) as the singular alignment target. The optimization objective is defined as minimizing the divergence between the state reached by the agent’s autonomous execution and the ground-truth $C_N$:
$$\mathcal{L}_{echo}(\theta) = \mathbb{E}_{(C_0, C_N) \sim \mathcal{D}_{echo}} [ \text{dist}(\text{Agent}_\theta(C_0), C_N) ],$$
where $\mathcal{D}_{echo}$ denotes the collection of interaction sequences retrieved from production service logs and $\text{Agent}_\theta(C_0)$ denotes the final state resulting from the entire sequence of state-transitioning steps performed by the agent. By optimizing for this objective, the agent learns to internalize the logic required to drive the environment toward the correct goal state. This allows the agent to generate the intended outcome autonomously, effectively replacing its previously flawed execution ($C_1$) with a self-driven trajectory that converges on the ground-truth $C_N$.

While the following section will demonstrate Echo in a coding environment, the $C_0	\rightarrow {C_1} \rightarrow {C_N}$ mechanism is essentially environment-agnostic, applicable to any agentic workflow where users inherently act as accountable stakeholders. 

\section{Practical Implementation: The Echo Pipeline in Auto-Completion}

\begin{figure}[t]
    \centering
    \includegraphics[width=0.9\textwidth]{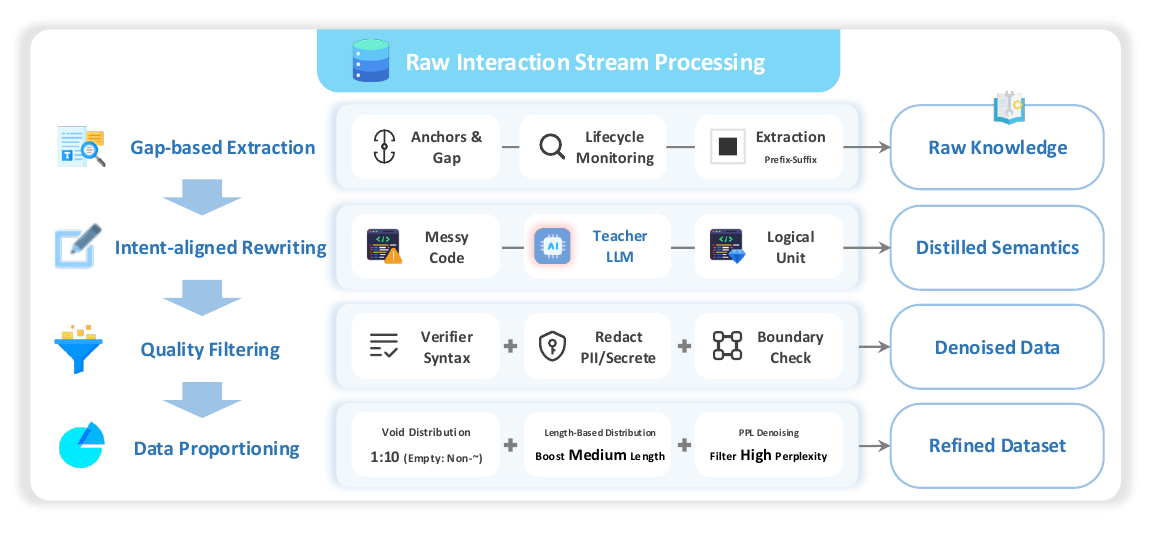}
    \caption{\textbf{Data Refinery Pipeline:} These raw residuals undergo a multi-stage rigorous filtering process. Through \textit{Gap-based Extraction} and \textit{Intent-aligned Rewriting} (guided by a Teacher LLM), noisy interactions are distilled into high-quality, intent-consistent training samples.}
    \label{fig:overview}
\end{figure}

In this section, we operationalize the Echo framework within the high-velocity domain of code auto-completion. We detail how the theoretical concept of "User-Driven Refinement" is translated into a rigorous engineering pipeline—transforming continuous editing streams into the extracted knowledge that enables perpetual agent evolution guided by accountable stakeholders.

\subsection{Context Tracking: Continuous Request Analysis}

Code auto-completion is formally defined as predicting a missing fragment $C$ given a Prefix ($P$) and a Suffix ($S$). A defining characteristic of this domain is that it operates as a high-frequency, continuous stream of requests. In the user's natural workflow, the interaction is seamless and constant. Whether the user accepts a recommendation or actively writes code manually, the IDE client continuously dispatches requests containing the updated Prefix and Suffix as the context evolves. This creates a dense, sequential record of the coding process where the context is updated with every keystroke~\citep{murali2023codecompose, ziegler2024measuring}.

Crucially, within this continuous stream, the agent frequently offers initial proposals ($C_1$). Whether the user directly accepts, partially modifies, or completely overrides these proposals, their ongoing keystrokes inherently constitute a continuous act of user-driven refinement. This mechanism allows us to deduce the true intent of the responsible stakeholder by analyzing the sequence of future requests. By observing the continuous stream, we can look ahead to see what code eventually stabilized in the user's editor~\citep{fried2022incoder, bavarian2022efficient}. This "future" content serves as the irrefutable ground truth for the "past" requests, enabling us to mine high-quality supervision signals directly from the natural evolution of the Prefix and Suffix without relying on explicit feedback buttons (e.g., binary user ratings like 'thumbs-up/down').

\subsection{Knowledge Mining: Gap-Based Extraction of $C_N$}

To accurately capture the user's verified intent ($C_N$), we track the lifecycle of the code gap established at the moment of the request.
\begin{itemize}[leftmargin=*]
    \item \textbf{Static Anchors and Gap Definition}: At any request moment $t_1$, the completion task is physically defined by two static context blocks: the Prefix ($P_1$) and the Suffix ($S_1$). The user's goal is to insert new code at the cursor position situated exactly between these two blocks. $$\text{State}_{t_1} = P_1 \oplus \text{<CURSOR>} \oplus S_1.$$ In this setup, both $P_1$ and $S_1$ serve as Static Anchors. They define the fixed "coordinates" of the user's intent, creating a stable gap that needs to be filled.
    \item \textbf{Lifecycle Monitoring via Dual Anchors}: As the user writes code (manually or via AI), the content of the file evolves. We track the status of this gap by monitoring the relative positions of the two anchors ($P_1$ and $S_1$) in the evolving document: (1) Active Filling: As long as both anchors $P_1$ and $S_1$ remain detected in the file and the cursor is positioned between them, the completion task is considered "In-Progress". The code growing between the anchors represents the user's ongoing solution. (2) Contextual Break (Termination): The mining session concludes when the Anchor relationship is broken. This occurs if the cursor moves to edit a different file or if the anchors themselves are modified or deleted, which signifies that the user has finished their work in this specific region.
    \item \textbf{Extraction}: Upon detecting contextual break, we extract the content accumulated between the two static anchors. This content represents the Commit State ($C_N$)—the final, verified code the user left behind to satisfy the requirement defined by $P_1$ and $S_1$, representing the ultimate destination of the user-driven refinement from any flawed initial proposal ($C_1$). 
\end{itemize}

\subsection{Signal Distillation: Intent-Aligned Truncation} 

Raw $C_N$ extracted via the contextual break logic represents the cumulative code written by the user before leaving the gap. Consequently, it often contains excessive future context—such as multiple subsequent lines or distinct logical blocks—that extends far beyond the single, atomic completion step the user actually needed at the specific trigger time. 

To bridge the gap between raw execution logs and high-quality training data, we introduce an automated refinement stage. Specifically, we deploy a high-capacity Instruction-Following LLM (acting as a "Teacher") to truncate and format the extracted content~\citep{gunasekar2023textbooks, raheja2023coedit}. It is imperative to emphasize that this Teacher LLM does not alter the logical correctness or the business logic verified by the accountable stakeholder. Instead, it strictly functions as an alignment tool, prompted to refine the raw $C_N$ with a strict focus on Truncation and Readability: 
\begin{itemize}[leftmargin=*]
    \item \textbf{Logical Scoping}: The raw code is segmented to capture the immediate logical unit (e.g., a single line of code, a function call, or even a fragmented segment) required by the user at the trigger time.
    \item \textbf{Human-Cognitive Alignment}: We truncate the code to ensure it fits within the user's immediate cognitive span. This is crucial because we treat the Refined $C_N$ as the training target—representing the exact suggestion the agent should have predicted at that moment. Consequently, it must be concise enough for a user to verify in seconds.
\end{itemize}

\subsection{Noise Reduction: Grounded Verification} 

To guarantee the validity and reliability of the mined experience data, we apply two distinct verification checks using an LLM-based verifier: 
\begin{itemize}[leftmargin=*]
    \item \textbf{Boundary Verification (Empty-Inference Correctness)}: In genuine outcome-oriented applications, the missing knowledge injected by the user is not always new code; it often manifests as "boundary awareness"—explicitly signaling when the agent should remain silent and yield control back to the human. A significant portion of the data pipeline focuses on Void Prediction. The verifier acts as a gatekeeper to ensure alignment between the context and the mined content. If the verifier determines that the context implies a "silence-required" state (e.g., deletions, navigation, or comments) but the mined data contains non-empty code (or conversely, if the verifier predicts code necessity but the data is empty), the sample is identified as a conflict and filtered out. This ensures the model is trained only on unambiguous signals, reinforcing clear boundaries for interference.
    \item \textbf{Robustness Verification}: We filter for code quality to prevent "poisoning" the model with bad user habits: (1) Correctness: Syntactic validity and variable consistency checks; (2) Safety: Filtering out hardcoded credentials, malicious patterns, or PII.
\end{itemize}

\subsection{Distribution Alignment: The Proportioning Strategy}

Instead of passively mirroring raw data distributions, we systematically adjust the training mixture guided by unified metrics from both offline benchmarks and online A/B tests. Our experimental optimization focuses on three primary dimensions:

\begin{itemize}[leftmargin=*]
    \item \textbf{Void Distribution}: We found that the optimal ratio of empty samples is not static but evolves with the scale of the dataset: (1) Baseline (50k samples): At a scale of 50k samples, a ratio of 1:10 (Empty : Non-Empty) achieves the optimal equilibrium. While "no-op" is common in reality, training on too many empty samples causes the model to become overly conservative (under-triggering). The 1:10 ratio provides sufficient negative constraints without dampening the model's helpfulness. (2) Scaling Trend: As the data volume increases, experimental results indicate that the proportion of non-empty samples should be increased. Since void prediction represents a "simple pattern" with low information density, over-exposure to these patterns—even at scale—can introduce optimization bias, causing the model to converge on "lazy" behaviors (preferring silence). Therefore, to sustain continuous improvement and handle complex corner cases, the density of non-empty (generative) samples should be fundamentally increased to prevent overfitting to the simplistic "do nothing" mode and ensure computational capacity is allocated to learning high-entropy reasoning tasks.
    \item \textbf{Length-Based Distribution}: We prioritized experiments on output length because length variance is a canonical characteristic of the auto-completion scenario. Experiments demonstrate that we should adjust the data proportioning to optimize information gain: (1) Suppress Short Data: We down-sample extremely short completions as they often represent trivial patterns (e.g., closing braces) that the model already masters; over-training here leads to "pattern hacking" rather than reasoning. (2) Boost Medium Data: We up-sample medium-length logical blocks, identifying this as the "sweet spot" where the density of the missing knowledge is highest (e.g., complex API usage or business logic). (3) Tail-End Pruning: Extremely long completions are down-sampled due to their statistical rarity and higher propensity for hallucinations.
    \item \textbf{PPL-Based Denoising}: We utilize Perplexity (PPL) as a proxy for "Surprisal". High-PPL user edits often indicate non-standard behaviors (e.g., pasting large distinct blocks). These are down-sampled to ensure the model learns generalized coding logic rather than idiosyncratic user noise~\citep{wenzek2020ccnet}.
\end{itemize}

\subsection{Optimization: Internalizing the Missing Knowledge} 

While the general Echo framework accommodates complex multi-step alignments (e.g., via Reinforcement Learning), the auto-completion scenario operates as a strictly single-turn generation task. In this specific domain, the distilled $C_N$ perfectly represents the exact, atomic string the agent should have output at the trigger time. Consequently, the generalized distance function naturally simplifies into direct Supervised Fine-Tuning (SFT). This allows us to directly model the probability of the refined $C_N$ conditioned on the initial context. The objective function is defined as:

$$\mathcal{L}_{Echo} = -\sum_{t=1}^{T} \log P_{\theta}(\text{Refined } C_N \mid \text{Prefix}, \text{Suffix}).$$

Using the directly optimized on the refined $C_N$, the model effectively internalizes the missing knowledge injected by the user. This allows the model to learn directly from human corrections in real-world environments, significantly enhancing its capabilities in practical applications.

\section{Evaluation}

In this section, we provide a comprehensive empirical evaluation of the Echo framework conducted within the real-world production environment. We demonstrate that learning from interaction experience not only significantly boosts performance, but also exhibits strong generalization and a clear scaling law.

\subsection{Evaluation Philosophy: Real-World Utility over Static Benchmarks}
Traditional academic evaluations of code models heavily rely on static benchmarks such as HumanEval or MBPP. However, as an industrial technical report focused on real-world utility, we deliberately de-emphasize these metrics in our primary evaluation. Static benchmarks excel at testing an agent's ability to solve isolated algorithmic puzzles, but they fundamentally fail to capture the heavy-tailed, messy reality of proprietary business logic, complex framework integrations, and the "unpublished" engineering knowledge required in production environments.

Optimizing strictly for public leaderboards often leads to "pattern hacking," which does not translate to actual productivity gains for developers. Therefore, our evaluation is anchored strictly in un-gamifiable, physical production metrics: Acceptance Rate (AR) and Generation Rate (GR). In the context of the Echo framework, these metrics are not merely product KPIs; they mathematically represent the agent's ability to autonomously hit the final correct state ($C_N$), thereby directly measuring the reduction in cognitive and physical costs required for user-driven refinement by accountable stakeholders. By measuring the exact frequency at which developers incorporate the model's proposals into their final commits during daily workflows, we provide a rigorous assessment of true intent alignment and the model's tangible value as a copilot.

\subsection{Experimental Setup}

\textbf{Data Source and Scale}: Our experiments were conducted using the CodeBuddy product. At the time of this study, the platform serves over 10,000 daily active users (DAU) working at Tencent Cloud, with an average of 273 completion requests per user per day. This high-frequency interaction environment provides a massive stream of data. To facilitate rapid experimental iteration and ensure fair comparison, we employed a fixed curated dataset of \textbf{50,000 experience samples} (Refined $C_N$) for our primary training.

\textbf{Base Model}: We used DeepSeek-Coder-6.7B~\citep{guo2024deepseek} as our base model, which was further optimized via continued pre-training~\citep{gururangan2020don} on 500B code tokens, leveraging its strong base capabilities in code generation.

\textbf{Metrics}: Following our evaluation philosophy, we evaluate the system using two primary production metrics~\citep{murali2024ai}:
\begin{itemize}[leftmargin=*]
\item \textbf{Acceptance Rate (AR):} The proportion of AI-suggested code blocks that are successfully adopted by the user.
\item \textbf{Generation Rate (GR):} The ratio of AI-generated code to the total code written by the user within a session, reflecting the model's overall contribution to productivity.
\end{itemize}

\subsection{Overall Performance}

\begin{figure}[t]
  \centering
  \begin{subfigure}[b]{0.46\linewidth}
    \centering
    \includegraphics[width=\linewidth]{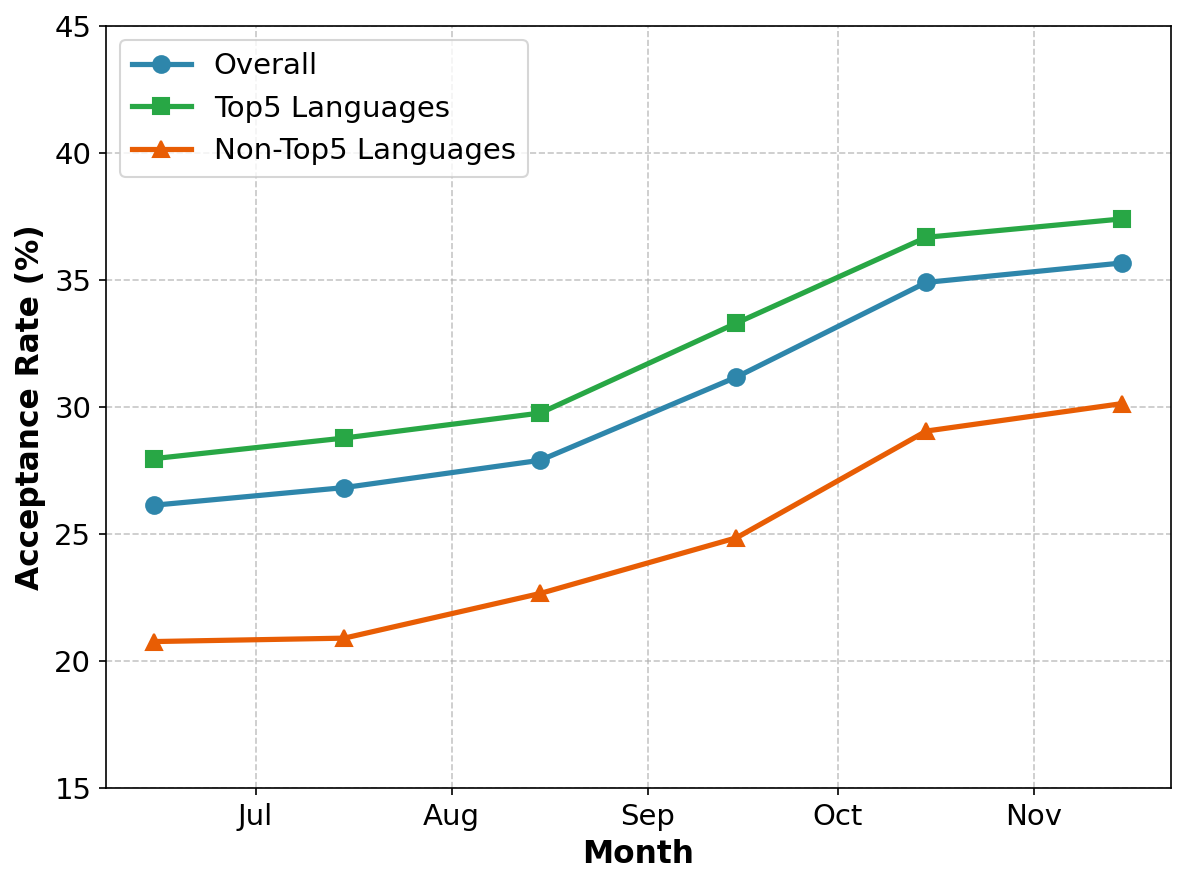}
    \caption{Monthly Acceptance Rate.}
    \label{fig:left}
  \end{subfigure}
  \hfill
  \begin{subfigure}[b]{0.46\linewidth}
    \centering
    \includegraphics[width=\linewidth]{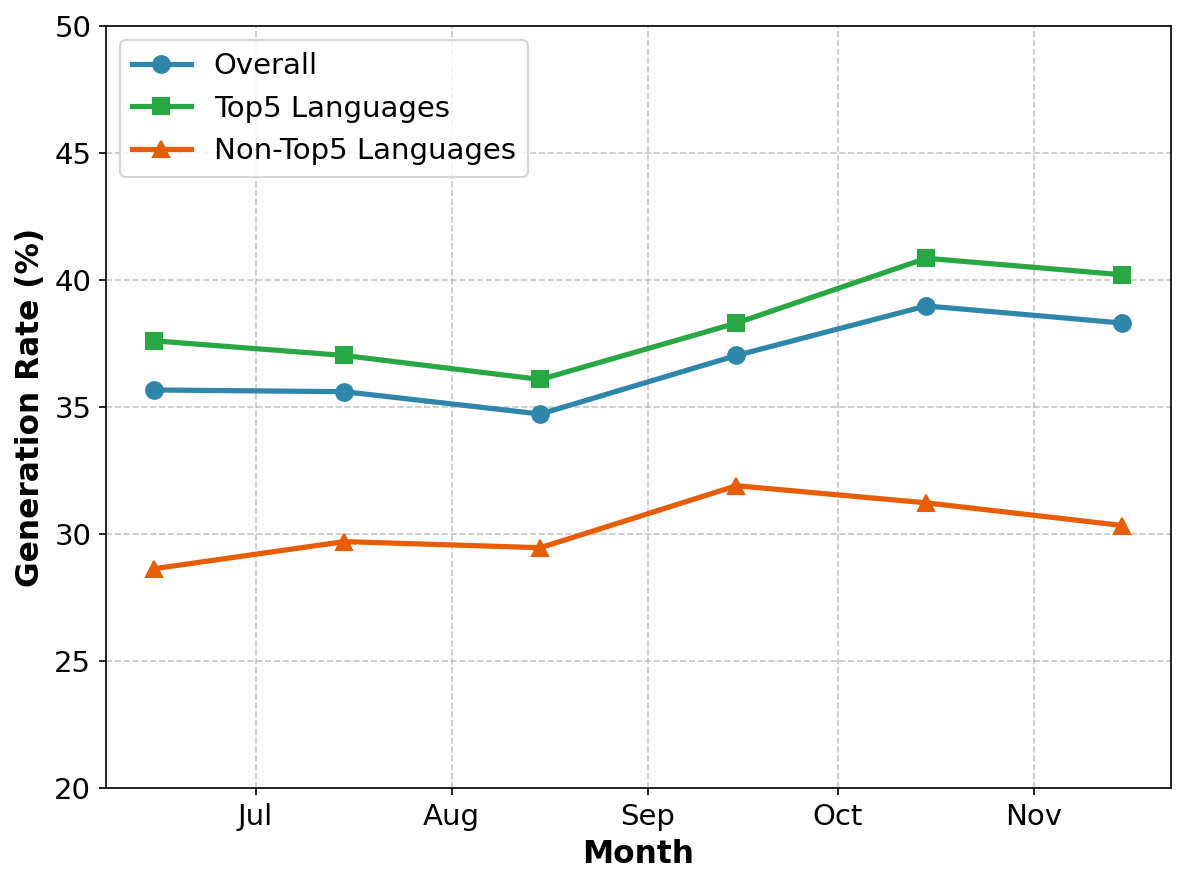}
    \caption{Monthly Generation Rate.}
    \label{fig:right}
  \end{subfigure}
  \caption{\textbf{Overall performance evolution.} The Echo pipeline breaks the static data ceiling, achieving a significant leap over the SFT baseline. Over the five-month period, AR improved by 10\% (reaching 35.7\%) and GR by 3.1\% (reaching 38.3\%).}
  \label{fig:overall_performance_five_months}
\end{figure}

Figure~\ref{fig:overall_performance_five_months} illustrates the performance trajectory over a five-month period within Tencent Cloud's internal environments. The starting point of the curve represents the baseline performance of the model trained purely on static "Human Data" (SFT). By iteratively refining the critical modules of the Echo pipeline—specifically Gap-based Extraction, Intent-Aligned Rewriting, Quality Verification, and Data Proportioning—we observed a substantial leap in performance: the Acceptance Rate (AR) surged from the static baseline of 25.7\% to 35.7\%, and the Generation Rate (GR) increased from 35.2\% to 38.3\%.

This trajectory highlights two critical insights. First, Echo effectively breaks the ceiling of static human data, achieving a significant performance leap beyond the SFT baseline. Second, the steady, monotonic increase is remarkable as it proves that stable model performance gains can be systematically achieved under the premise of a reasonably structured engineering pipeline organization, validating Echo as a robust framework for experience learning. Physically, this upward trend signifies that the agent is successfully internalizing the extracted missing knowledge, enabling it to output the ground-truth $C_N$ autonomously and significantly reducing the necessity for human intervention.

\subsection{Generalization to External Environments}

\begin{table}[t]
\centering
\small
\setlength{\tabcolsep}{12pt}
\renewcommand{\arraystretch}{1.2}
\begin{tabular}{lcccc}
\toprule
Model &
\multicolumn{2}{c}{\textbf{Codewise-7B-SFT}} &
\multicolumn{2}{c}{\textbf{Codewise-7B-Echo}} \\
\hline
Metrics
& \textbf{AR} & \textbf{GR}
& \textbf{AR} & \textbf{GR} \\
\midrule
All                    & 25.01 & 36.05 & 37.55 & 40.58 \\
Top-6 Languages        & 26.82 & 37.14 & 39.70 & 42.93 \\
Vue                    & 25.22 & 37.22 & 38.74 & 40.15 \\
Python                 & 24.80 & 41.30 & 41.78 & 46.83 \\
Java                   & 28.59 & 31.84 & 38.26 & 38.98 \\
JavaScript/TypeScript  & 27.29 & 41.43 & 40.67 & 46.46 \\
C/C++                  & 25.30 & 47.67 & 39.35 & 49.03 \\
Go                     & 31.08 & 37.35 & 42.91 & 44.11 \\
Other                  & 20.53 & 32.96 & 32.00 & 34.69 \\
\bottomrule
\end{tabular}
\vspace{1em}
\caption{
Online code-completion metrics across language segments for an external-user deployment.
\textbf{Length} is the average number of generated characters per suggestion (chars/completion).
}
\label{tab:Generalization_to_External_Environments}
\end{table}

A key concern when training on data derived from a specific user population is the risk of overfitting to that group's unique behaviors, which typically leads to mediocre performance when applied to external user groups. To test the generalization capability of Echo, we deployed the model to external users. The results are shown in Table~\ref{tab:Generalization_to_External_Environments}.Despite the entirely different distribution of coding styles and project contexts, we observed consistent and substantial improvements: the overall Acceptance Rate increased from 25.01\% to 37.55\%, and the Generation Rate rose from 36.05\% to 40.58\%. This trend of stable improvement was evident across all major programming languages. 

It is worth noting that while there are slight variances in statistical baselines due to differences in product instrumentation (statistical caliber) between the internal and external versions, the relative performance gains within the same product environment are objective and significant. This evidence strongly suggests that the $C_N$ extraction process captures fundamental intent alignment capabilities that are robust across different organizations and environments. 

\subsection{Scaling Effect of Experience Data}

\begin{figure}[h]
  \centering
  \includegraphics[width=0.92\linewidth]{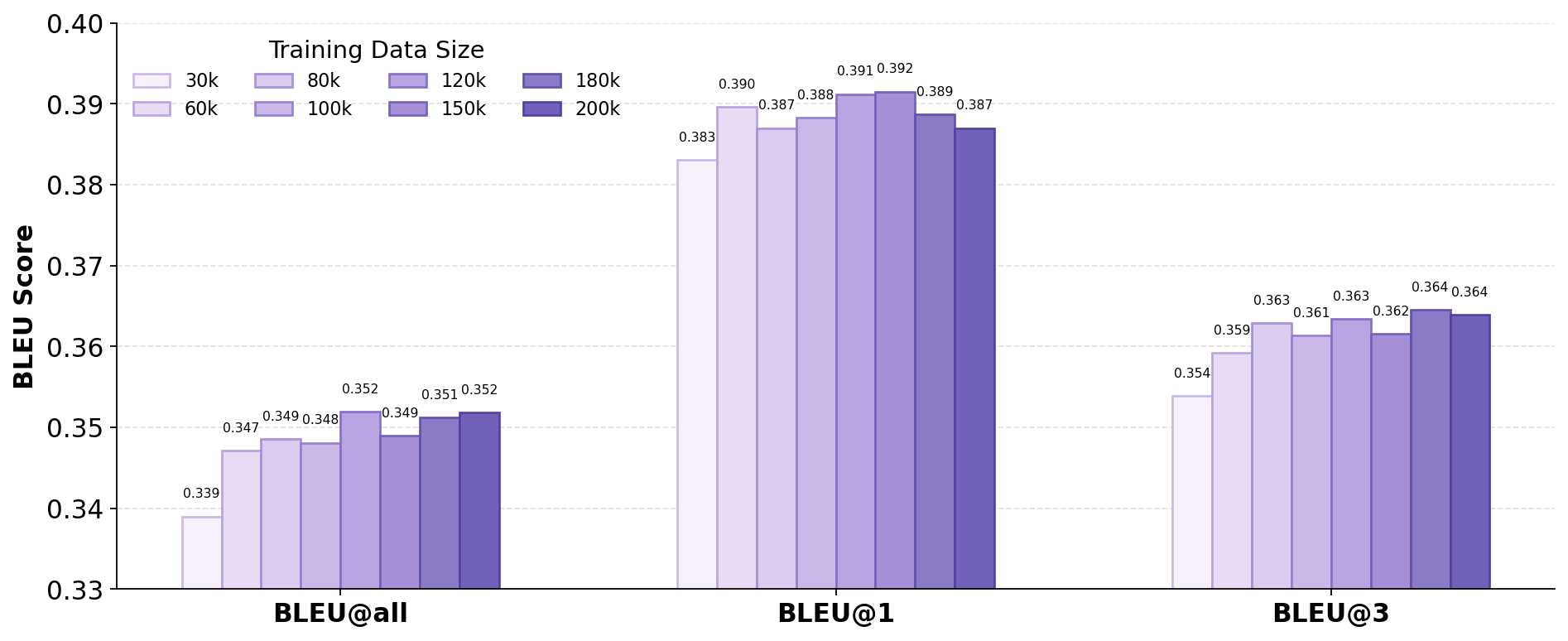}
  \caption{\textbf{Scaling Effect of Experience Data.} We vary the training data volume from 30k to 200k samples in the Python domain and evaluate performance using BLEU. The results show a clear scaling effect as experience data increases.}
  \label{fig:scaling-effect-bleu}
\end{figure}



We investigated the relationship between data scale and model capability. Due to constraints on GPU training resources, we restricted this specific scaling analysis to the Python domain, varying the training data volume from 30,000 to 200,000 samples. Performance was evaluated primarily through offline metrics, utilizing BLEU scores~\citep{papineni2002bleu, ren2020codebleu} on a proprietary benchmark dataset constructed from real-world usage scenarios to ensure rigorous assessment.

The experimental results in Figure~\ref{fig:scaling-effect-bleu} exhibit a clear \textbf{Scaling Effect}, demonstrating that model performance improves effectively as the volume of experience data increases. Although we observed certain outliers at specific data points, these are attributable to inherent instabilities in the training process rather than a saturation of the learning signal.

This observation carries profound implications. First, it proves the long-term potential of the Echo paradigm; the absence of a performance plateau indicates that we can drive continuous improvement simply by scaling the accumulation of interaction data. Second, and perhaps more critically, in an era where Large Language Models are increasingly bottlenecked by the exhaustion of static human data, the scalability of the Echo framework offers a paradigm shift. By inherently coupling the model's learning process with the continuous, responsibility-driven interventions of accountable stakeholders, Echo establishes an inexhaustible data engine. This ensures a sustainable path for perpetual agent evolution, bounded only by the scale of real-world application rather than the finite limits of historical archives.

\section{Discussion}

The results of Project Echo extend beyond incremental performance gains. They illuminate a new paradigm for agent evolution that challenges the current heavy reliance on static pre-training and imitation learning~\citep{kaplan2020scaling, hoffmann2022training}.

\subsection{The Information Entropy of Correction Paths}

Unlike traditional paradigms (e.g., SFT or RLHF) that optimize against "static targets"~\citep{ouyang2022training}, Echo learns directly from the human correction outcomes. The entropy contained in a user correcting a model's flawed proposal ($C_1$) to the final commit ($C_N$) is significantly higher than that of a standalone code snippet. This continuous act of user-driven refinement explicitly exposes the ignorance boundary of the model's current world model. By actively injecting the missing knowledge to bridge the gap between the initial belief and the verified environmental requirement, the accountable stakeholder provides a uniquely dense supervision signal. This targeted learning is what allows Echo to break the performance ceiling of human-data-only models and move toward out-of-distribution robustness.

\subsection{The Strategic Moat of Long-Tail Data}

As foundational models increasingly converge in base capabilities, the frontier of AI competition is shifting toward the ability to capture and internalize authentic, long-tail data. The Echo paradigm demonstrates that widespread product deployment is not merely the commercial end-goal of a model, but a critical mechanism for long-tail data exposure~\citep{halevy2009unreasonable, villalobos2024position}. By embedding the agent directly into real-world professional contexts, the product serves as a continuous sensor, actively exposing the model to a vast array of "long-tail" requirements and proprietary business logic that are entirely absent from internet corpora. 

This creates a self-reinforcing Strategic Moat: as the model handles complex project-specific logic more effectively, it attracts a larger base of accountable stakeholders, which in turn accelerates the flywheel of our inexhaustible data engine. In this regime, the leading agent will not necessarily be the one pre-trained on the largest static corpus, but the one that most efficiently transforms its real-world exposure into grounded model intelligence. The ability to capture and learn from these unique, long-tail distributions becomes the primary barrier to entry for competitors.

\subsection{Beyond the Pre-training Paradigm: Tapping into the Unpublished World}

Perhaps the most profound implication of Echo is its potential to surpass the contribution of pre-training to model capability. Traditional pre-training data consists of human-published content, which represents only a small, often biased fraction of the total human intellectual output. The vast majority of human problem-solving logic and proprietary expert knowledge remains \textbf{"unpublished"}---locked within private workflows, internal project logic, and real-time decision-making processes~\citep{lee2022deduplicating}.

As agents become ubiquitous in professional environments, the Echo framework enables the capture of this massive, untapped data asset. Unlike the "digital ink" of the web, this experience data is: (1) Execution-Aligned: Because user refinements ($C_N$) are driven by the need for code to function in a real-world production environment, the data is naturally grounded in the deterministic constraints of production and execution reality, rather than just linguistic patterns. (2) High-Fidelity: It represents the actual "working logic" of experts rather than the "curated narratives" found in public publications. We hypothesize that as experience data scales, its role may evolve from a fine-tuning signal to a primary driver of emergent intelligence, potentially making traditional pre-training on public text less central to the development of superhuman agents.

\subsection{Limitations and Data Governance}

While the potential is vast, extraction from live sessions requires rigorous de-identification and anonymization to protect user privacy and proprietary logic. Furthermore, the framework must systematically address the noise-quality trade-off.

As the agent's baseline capabilities grow, a critical challenge arises: ensuring that the extracted $C_N$ consistently represents the optimal "truth" rather than a developer's temporary "compromise" (e.g., quick hacks or dirty workarounds). Future iterations must evolve beyond simple heuristic filtering to construct an Automated Alignment Oracle—a multi-dimensional quality evaluation system that can rigorously assess whether a user's refinement truly adheres to the highest standards of execution reality. Overcoming this data governance challenge will be the key to unlocking the next order of magnitude in agentic intelligence.

\section{Conclusion and Future Work}

In this paper, we introduced \textbf{Project Echo}, a practical framework designed to overcome the imitation bottleneck of static human-generated datasets. By redefining agentic applications as high-frequency sensors actively capturing long-tail data and execution reality, and modeling the interactive refinement trajectory from the initial proposal ($C_1$) to the user-committed final version ($C_N$), we demonstrated a scalable path for agent self-evolution. Our large-scale validation in the CodeBuddy environment proves that Echo can break the performance ceiling of traditional SFT paradigms, increasing the code acceptance rate from 26\% to 36\%. More importantly, the robust zero-shot generalization to external users suggests that Echo captures fundamental alignment signals that transcend specific organizational distributions. Ultimately, by internalizing the missing knowledge and drastically reducing the cognitive and physical costs of user-driven refinement for accountable stakeholders, Project Echo establishes a robust, inexhaustible data engine for continuous agentic evolution.

\subsection{Future Work}

While Project Echo marks a successful step toward operationalizing experience data, it represents only the beginning of this paradigm shift. We identify three critical dimensions for future exploration:

\begin{itemize}[leftmargin=*]
    \item \textbf{From Completion to Complex Agents:} Code completion, while high-frequency, is a relatively restricted task. The most immediate frontier for the Echo paradigm is its application to code agents capable of multi-step reasoning and tool use. In these scenarios, the "echo" ($C_N$) will no longer be a single static code snippet, but rather a verified execution trajectory or a complex environment state comprising sequential file-system changes. Extending our $C_1 \to C_N$ modeling to capture and compress these long-horizon planning trajectories will be essential for creating truly autonomous experts.
    
    \item \textbf{Extending the Temporal Horizon of Experience:} Current Echo implementation focuses on relatively short interaction streams, typically spanning minutes to hours. However, many real-world engineering tasks have lifecycles spanning days or weeks. In such long-span contexts, identifying which specific agent proposal contributed to the final "Commit State" becomes a significant credit-assignment challenge. We plan to investigate temporal anchoring algorithms that can extract high-fidelity supervision signals across discontinuous and long-duration sessions.
    
    \item \textbf{Learning Beyond Objective Ground Truth:} Echo currently relies on the user-driven refinement mechanism in domains where the user is compelled to produce a "correct" functional answer (e.g., code that compiles and runs). However, in many creative or subjective domains, a single "correct" objective ground truth does not exist. We aim to explore how the Echo paradigm can be adapted to scenarios where the objective is "satisfaction" rather than absolute "verification." In these contexts, the final adoption by the accountable stakeholder inherently constitutes the subjective execution reality. Adapting Echo to these environments will potentially merge interactive experience learning with latent preference modeling. 
\end{itemize}

Through these efforts, we hope to advance Project Echo from a specialized tool for developers into a general framework for agents that continuously learn, adapt, and excel through their lived experience in the human world.

\section*{Acknowledgements}
The author would like to express sincere gratitude to Yi Liu for the valuable guidance, insightful discussions, and continuous support throughout the development of this project. Special thanks go to Zhezheng Hao and Tianfu Wang for their excellent assistance in preparing the figures and polishing the manuscript.

\bibliographystyle{abbrvnat}
\bibliography{main}

\appendix

\end{document}